\documentclass[conference]{IEEEtran}
\IEEEoverridecommandlockouts
\usepackage{cite}
\usepackage{amsmath,amssymb,amsfonts}
\usepackage{algorithmic}
\usepackage{graphicx}
\usepackage{textcomp}
\usepackage{xcolor}
\def\BibTeX{{\rm B\kern-.05em{\sc i\kern-.025em b}\kern-.08em
    T\kern-.1667em\lower.7ex\hbox{E}\kern-.125emX}}
    
\usepackage{xspace}
\usepackage{multirow}
\usepackage{subcaption}
\usepackage[hidelinks]{hyperref}
\usepackage{xurl}
\newcommand{\aicab}{\textsc{AIC-AB Net}\xspace}

\begin{document}

\title{\aicab: A Neural Network for Image Captioning with Spatial Attention and Text Attributes}


\author{
\IEEEauthorblockN{Guoyun Tu}
\IEEEauthorblockA{\textit{Department of Computer Science} \\
\textit{KTH Royal Institute of Technology}\\
Stockholm, Sweden \\
guoyun@kth.se}
\and
\IEEEauthorblockN{Ying Liu}
\IEEEauthorblockA{\textit{Norna}\\
Stockholm, Sweden \\
ying.liu@norna.ai}
\and
\IEEEauthorblockN{Vladimir Vlassov}
\IEEEauthorblockA{\textit{Department of Computer Science} \\
\textit{KTH Royal Institute of Technology}\\
Stockholm, Sweden \\
vladv@kth.se}
}

\maketitle

\begin{abstract}
Image captioning is a significant field across computer vision and natural language processing. We propose and present \aicab, a novel Attribute-Information-Combined Attention-Based Network that combines spatial attention architecture and text attributes in an encoder-decoder. 
For caption generation, adaptive spatial attention determines which image region best represents the image and whether to attend to the visual features or the visual sentinel. Text attribute information is synchronously fed into the decoder to help image recognition and reduce uncertainty. 
We have tested and evaluated our \aicab on the MS COCO dataset and a new proposed Fashion dataset. The Fashion dataset is employed as a benchmark of single-object images. The results show the superior performance of the proposed model compared to the state-of-the-art baseline and ablated models on both the images from MSCOCO and our single-object images. Our \aicab outperforms the baseline adaptive attention network by 0.017 (CIDEr score) on the MS COCO dataset and 0.095 (CIDEr score) on the Fashion dataset.
\end{abstract}

\begin{IEEEkeywords}
image captioning, neural networks, spatial attention, text attributes
\end{IEEEkeywords}

Regular Research Paper

\section{Introduction}\label{ch:introduction}
The significant growth in web images has brought plenty of opportunities for computational understanding of images. Automatic image captioning is crucial for many applications, including image searching, categorizing, and indexing, and it has attracted attention from academia and industry. One can split the image captioning task into two parts (1) image recognition to detect and recognize objects in an image and (2) caption generation to summarize the extracted information and put it into text that humans understand. 

Significant successes have been achieved in the problem of image captioning using Deep Learning (DL). Many works on image captioning have applied DL methods to images containing multiple objects and rich contextual information. As a result, various image captioning methods have been proposed, such as Visual space-based model~\cite{chen2015}, multimodal space-based model~\cite{kiros14}, dense captioning~\cite{JohnsonKL15}, whole scene-based model, encoder-decoder architecture-based model~\cite{VinyalsTBE14}, compositional architecture-based model~\cite{fang2015}, attention-based model~\cite{xu2015}, semantic concept-based model~\cite{Karpathy2017}, stylized captions~\cite{gan2017}. We address the following two problems (1) generating captioning on single-object images; (2) combining semantic text attributes and adaptive attention. 

The first blind spot of the previous studies is that they focused on general multi-object images, whereas generating captioning on single-object images is barely studied. Two features distinguish single-object images from general images. First, single-object images contain more small details, thus requiring a higher recognition resolution. Second, the generated descriptions include more adjectives and nouns. In this work, we apply DL models on a fashion dataset of 144,422 images from 24,649 products. This dataset is used as a benchmark of single-object images. Each image has only one fashion item, and its caption describes that item, including its category, color, texture, and other details.

Secondly, previous DL approaches either boost image captioning with semantic concept~\cite{fang2015,ting2016} or make use of attention encoder-decoder framework~\cite{xu2015,Lu2017}, i.e., two frameworks are used separately and cannot inform each other. We propose a novel attribute-image-combined attention-based neural network architecture (\aicab\footnote{\url{https://github.com/guoyuntu/Image-Captioning-On-General-Data-And-Fashion-Data}}) based on the adaptive attention network~\cite{Lu2017}. It combines the semantic concept-based architecture with the spatial attention-based architecture. 
In \aicab, the text attributes are fed into each step of the LSTM decoder as an additional input when generating the captions. The attributes are obtained by an auxiliary CNN classifier.

The major contributions of our work are as follows. 
\begin{enumerate}
    \item We propose an Attribute-Image-Combined Attention-Based Network (\aicab) that combines the adaptive attention architecture and text attributes in an encoder-decoder framework and, as a consequence, improves the accuracy of image captioning compared to state-of-the-art alternatives.
    \item We evaluate \aicab and several other DL models on a single-object dataset, the Fashion dataset, containing 144,422 images from 24,649 products.
\end{enumerate}

\section{Related work}\label{sec:related}

Prior works, e.g.,~\cite{VinyalsTBE14,Wang2016,Jia2015,Wuqi2016}, use DL encoder-decoder methods, attention-based and semantic concept-based DL models for automated image captioning. 

\textbf{Encoder-Decoder for Image Captioning}. The existing caption generation methods include template-based image captioning ~\cite{Farhadi2010}, retrieval-based image captioning ~\cite{Yunchao2014}, and language model caption generation ~\cite{KirosSZ14,xu2015,you2016}.
Most methods~\cite{mao2017,Wang2016,Jia2015,zhao2018} use Deep Learning. 
An encoder-decoder, 
a popular approach to tackling language tasks, such as machine translation, can be used for image captioning to encode visual information and decode it in a natural language. A 
network of this category extracts global image features from the hidden activations of a CNN and feeds them into an LSTM to generate a caption as a sequence of words; one word at each step depends on a context vector, the previous hidden state, and the previously generated words~\cite{KirosSZ14}.

\textbf{Attention-based Networks}. Following the trends to use the encoder-decoder architecture on image captioning, methods based on attention mechanisms~\cite{Huang2020,xu2015,Lu2017} have been increasingly popular as they provide computer vision algorithms with the ability to know where to look. Instead of considering the image as a whole scene, an attention-based network dynamically focuses on various parts of the input image while generating captions.

An adaptive attention network is an encoder-decoder-based approach for image captioning. The decoding stage is split into two parts. 
First, the Spatial Attention Network outputs a context vector $c_t$ that depends on the feature map $V$ extracted from the encoder and the hidden state $h_t$ of the LSTM decoder. It could be considered as the attention map. The second part is Visual Sentinel, which can fall back on when it chooses not to attend to the image. This visual sentinel $s_t$ is dependent on the input $x_t$, the hidden state $h_{t-1}$, and the memory cell $m_t$. Then, the new adaptive context vector is modeled as $\mathbf{c}_t = \sum_{i=1}^k{\alpha}_{ti}{\upsilon }_{ti}$. $\mathbf{c}_t$ and $h_t$ 
determine the conditional probability for each time step of LSTM.

\textbf{Semantic concept-based Models}. The idea of semantic concept-based models is to extract a set of semantic concept proposals. These concepts, combined with visual features and hidden states, are used to generate the captions in the decoding stage. Karpathy et al.~\cite{Karpathy2017} proposed a model, in which dependency tree relations 
are applied in training to map the sentence segments with the image regions with a fixed window context. Wu et al.~\cite{Wuqi2016} proposed a network, including high-level semantic concepts explicitly. 
It adds an intermediate attribute prediction layer in an encoder-decoder framework to extract from images attributes used to generate semantically rich image captions.
The proposed technique in this paper is partially inspired by~\cite{ting2016}, where Ting et al. suggested that the high-level attributes are more semantically rich and easily translated into understandable human sentences. Moreover, the best method is to feed attribute representations and visual features as a joint input to LSTM at each time step. 
The prior works are limited to using 
only one architecture, either semantic concept-based or spatial attention-based. 
We propose \aicab that combines both structures so that they can communicate with each other and, as a consequence, improve the accuracy of image captioning.

\section{\aicab: Attribute-Image-Combined Attention-Based Network}\label{sec:aicabnet}
\begin{figure*}
\centering
\includegraphics[width=\textwidth]{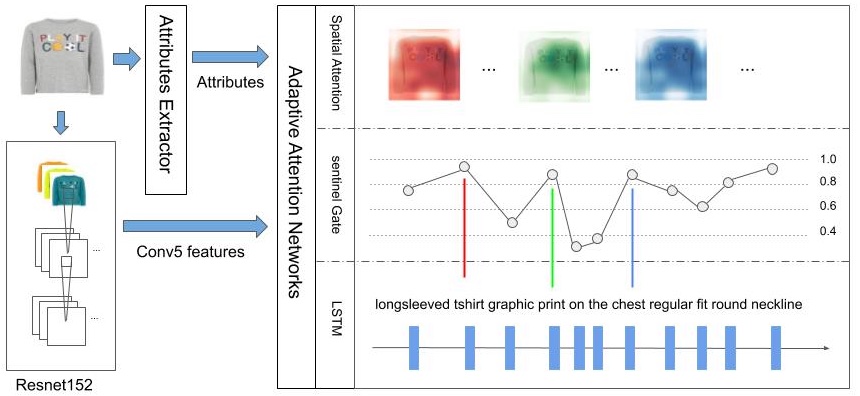}
\caption{Overview of \aicab. It extracts from the images the visual features and the attribute information. The former pass to the adaptive attention architecture~\cite{Lu2017}, the latter are fed into the LSTM decoder at every time step.}
\label{fig:model_overview}
\end{figure*}

We present our Attribute-image-combined Attention-based Network (\aicab), a novel encoder-decoder neural framework for image captioning. \aicab is an end-to-end network that tackles image captioning in the meantime, generates the attention map of the image. Fig.~\ref{fig:model_overview} shows the network architecture. We extract image features using pre-trained ResNet-152~\cite{he2016}, which implements residual learning units to alleviate the degradation of deep neural networks. We freeze the first six layers and take the last convolutional layer as visual features. We believe the features extracted retain both object and interaction information from the images.

Formally, let us denote the whole dataset as $\mathfrak{D} = \{(\mathbf{X}_i,\mathbf{y}_i\} (i=1,2,...,N)$ where $\mathbf{X}_i$ denotes the 
$i$-th image and $\mathbf{y}_i = (y_1,y_2,...,y_t)$ denotes its caption label as a sequence of words. In an encoder-decoder framework, LSTM network plays the role of decoder and each conditional probability is modeled as:
\begin{equation}
\setlength{\abovedisplayskip}{3pt}
\setlength{\belowdisplayskip}{3pt}
    \sum_{t=1}^{L}\log p(y_t \vert y_1,y_2,...y_{t-1},\mathbf{X}) = f(\mathbf{h}_t,\mathbf{c}_t)
    \label{eq3}
\end{equation}
where $f$ is a nonlinear function that outputs the probability of $y_t$. $\mathbf{c}_t$ is the visual context vector at time step $t$ extracted from image $\mathbf{X}$. $\mathbf{h}_t$ denotes the hidden state at $t$. For LSTM, $\mathbf{h}_t$ could be modeled as:
\begin{equation}
\setlength{\abovedisplayskip}{3pt}
\setlength{\belowdisplayskip}{3pt}
    \mathbf{h}_t = LSTM(\mathbf{x}_t,\mathbf{h}_{t-1},\mathbf{m}_{t-1})
    \label{eq4}
\end{equation}
where $\mathbf{x}_t$ is the input feature map, $\mathbf{m}_{t-1}$ is the memory cell vector at 
$t-1$.

\subsection{Text Attribute Extractor}
The first step in adding the text attributes into the LSTM decoder is to extract a set of words that are possible to attend in the image’s description. These words may belong to the following parts of speech, including nouns and adjectives. As suggested by~\cite{fang2015}, we build the vocabulary $V$ using the 1000 most common words of the training captions. Given a vocabulary of attributes, the next step is to detect these words from images. We train the text attribute extractors using a CNN-based model. An image passes the pre-trained VGG-16 model, and we express the Cov5 layer as the input feature map, which is fed into a 2-layer CNN following with one fully connected layer. The possibility $p_i^w$ of image $x_i$ containing word $w$ is computed by a sigmoid layer:
\begin{equation}
\setlength{\abovedisplayskip}{3pt}
\setlength{\belowdisplayskip}{3pt}
    p_t^w = \frac{1}{1+\exp(-(v_t^w\phi(b_{i})+u_w))}
    \label{eq12}
\end{equation}
here $\phi(b_i)$ is the fully connected representation for image $b_i$, $v_t^w$ and $u_w$ are the associated weights and bias with word $w$.

Due to the highly imbalanced ratio of the positive labels (5 words per image) vs. negative, The loss function used for training the detector is 
\begin{equation}
\setlength{\abovedisplayskip}{3pt}
\setlength{\belowdisplayskip}{3pt}
    \mathfrak{L}_i^C = -{\beta}_p(p(x_i)\log(q_i))+\beta_n(1-p(x_i))(\log(1-q(x_i)))
    \label{eq13}
\end{equation}
where $\beta_p$ and $\beta_n$ are class weights assigned for giving higher penalty over false positive predictions. Due to the very unbalanced labeling strategy, 
$\beta_p = 100 \dot \beta_n$.

\subsection{Attribute-combined Model}

\begin{figure}[b]
\centering
\includegraphics[width=\columnwidth]{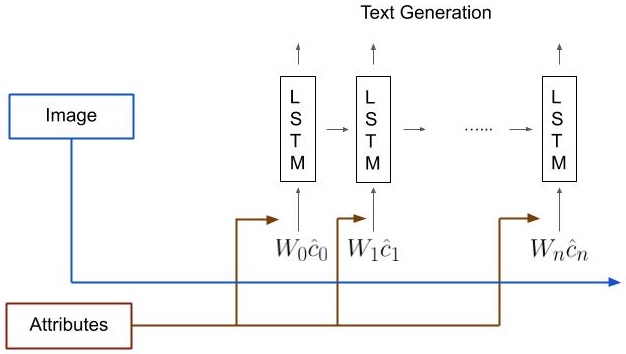}
\caption{A simplified diagram of our attribute-image-combined network (\aicab).}
\label{fig:attributecombinedmodel}
\end{figure}

With injecting the high-level attributes into the adaptive attention framework, we propose our \aicab (Fig.~\ref{fig:model_overview}). In our model, the decoder is modified by additionally integrating visual information and high-level attributes. As Fig.~\ref{fig:attributecombinedmodel} shows, the encoded image features are fed from the start of the LSTM and text attributes are fed into each time step. Accordingly, given the attribute representation $\mathbf{A}$ the calculation of the hidden state in each time step is converted from Eq.~\eqref{eq4} to:
\begin{equation}
\setlength{\abovedisplayskip}{3pt}
\setlength{\belowdisplayskip}{3pt}
    \mathbf{h}_t = LSTM(\mathbf{x}_t, \mathbf{A},\mathbf{h}_{t-1},\mathbf{m}_{t-1})
    \label{eq14}
\end{equation}

\begin{figure}
\centering
\includegraphics[width=\columnwidth]{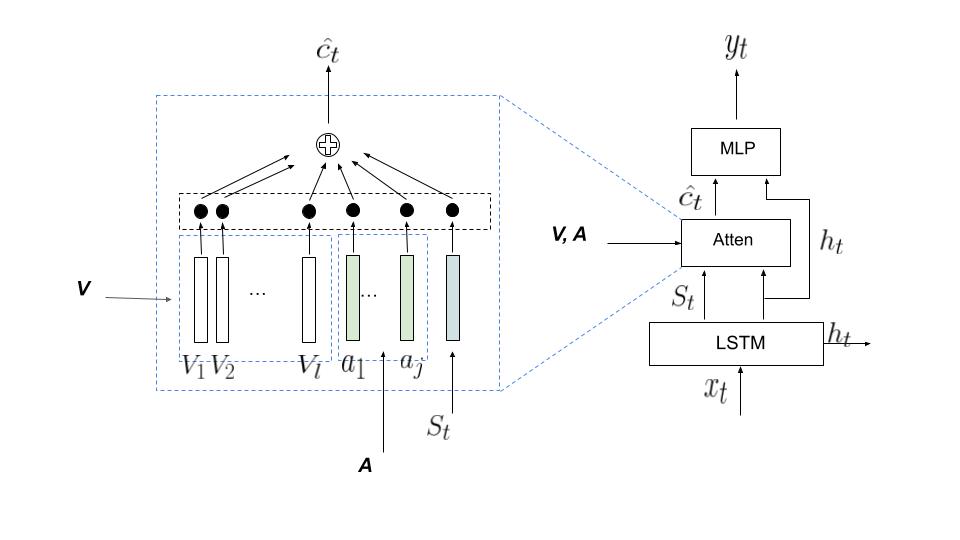}
\caption{An illustration of \aicab generating the $t$-th word $y_t$.
The input is the encoded image features $V$ and attribute information $A$.}
\label{fig:adaptive_attention}
\end{figure}

Adding text attributes to the original adaptive attention framework. The architecture of \aicab at one time step is demonstrated in Fig.~\ref{fig:adaptive_attention}. The probability over the vocabulary at time step $t$ can be computed as:
\begin{equation}
\setlength{\abovedisplayskip}{3pt}
\setlength{\belowdisplayskip}{3pt}
    \mathbf{p}_t = \mathrm{softmax}(\mathbf{W}_p(\hat{\mathbf{c}}_t+\mathbf{h}_t))
    \label{eq11}
\end{equation}
where $\mathbf{W}_p$ is weight parameter to be learnt.

\section{Evaluation Setup}\label{sec:experiments}
To evaluate the proposed \aicab, we have conducted evaluation experiments using multi-object and single-object image datasets and have compared the performance of our \aicab with competing baselines.

\subsection{Datasets and Preprocessing}\label{dataset}
In our evaluation, we have used two image datasets, the MS COCO dataset~\cite{linyi2014} and a fashion image dataset. 

\textbf{MS COCO Dataset}~\cite{linyi2014} contains 328K images with a total of 2.5M 
object instances. For the image captioning task, five 
caption descriptions labeled for each image are used as ground truth. 
We use the MS COCO dataset to compare the performance of 
\aicab with a state-of-the-art work
~\cite{Lu2017}.

\textbf{Fashion Dataset} is our single-object fashion image dataset scraped in open websites from different fashion vendors, including Uniqlo, Toteme-Studio, Bestseller, Drykorn, Jlindeberg, Joseph-fashion, Marc-o-polo, Rodebjer, Tigerofsweden, and Vince. The raw data contains 1,511,916 images from 194,453 fashion products. Each data includes a text label consisting of various amounts of sentences. 
Before the image captioning task, we have conducted a data cleaning task to remove images with invalid or unrelated captions.
The cleaned dataset includes 144,422 images from 24,649 products. 
Each data is labeled with only one caption description sentence. 
Products are allowed to map to the same caption, and there are 10,091 unique captions in this dataset. 

\textbf{Preprocessing}. We apply the same split for COCO and Fashion datasets: 70\% of the data for training, 15\% for validation, and 15\% for testing. 
We resize all the images used in experiments to $224\times224$ with bilinear interpolation. We also create two variations for the Fashion dataset. The one-vendor condition focuses on the largest product vendor, Bestseller. Further called the \textbf{Fashion Bestseller dataset}, this subset contains 89,756 images from 19,385 products. The amount of unique captions is 8,448. The second condition employs all images in the dataset, which we further call the \textbf{Fashion 9 vendors dataset}. The reason behind it is that we found the same vendor usually describes their product in similar text form and style. For all three datasets, COCO, Fashion Bestseller, and Fashion 9 vendors, we automatically generate five attributes for each image with the following method to train the attribute extractor. First, we build an attribute vocabulary comprising 1000 most common words (nouns and adjectives) from the caption text. Then, we choose five words in the caption which occur in the vocabulary as attributes for each data.

\subsection{Hyperparameters; Baseline and Ablated Models}
\textbf{Text Attribute Extractor}. The convolution layer of the text attribute extractor has kernel size $5 \times 5$ and stride $1 \times 1$. The max pooling layer has kernel size $8 \times 8$ and stride $0$. We train the text attribute extractor for 10 epochs.

\textbf{\aicab network}. In the decoding stage, words in captions are embedded into 255-dimension vectors, and words of attributes are embedded into 51-dimension vectors using the default word embedding function provided by PyTorch. The hidden size is set as $512$. Adam optimizer with learning rate decay is employed to train the model. The parameters is set as: $\alpha = 0.8$, $\beta = 0.999$, $learning\_rate = 4e-4$. The decay of learning rate is modeled as:
\begin{equation}
\setlength{\abovedisplayskip}{3pt}
\setlength{\belowdisplayskip}{3pt}
    l_r^{E+1} = l_r^E * {0.5}^{\frac{E-20}{50}}, E > 20
    \label{eq15}
\end{equation}
where $l_r^E$ 
is the learning rate in epoch $E$. We train the extractor for 50 epochs.

 We compare 
 \aicab to the following baseline and ablated models.
\begin{enumerate}[1.]
\item \textbf{Adaptive}: the state-of-the-art method, Adaptive~\cite{Lu2017}. Note that our model without the text attributes (i.e., being ablated) is the same as Adaptive.


\item \textbf{The Vanilla Encoder-Decoder (Vanilla-ED)}: an ablated \aicab, where we remove the adaptive attention architecture and attribute information while keeping a CNN-based encoder and LSTM-based decoder.

\item \textbf{The Text Attributes Only (Attr-Only)}: a second ablated \aicab, where we feed the attribute information into the LSTM decoder but remove the attention architecture.
\end{enumerate}

\section{Results and Discussion}\label{sec:results}

\begin{table*}
    \caption{\label{tab:image_captioning_all_three} Evaluation Scores of Image Captioning}
    \centering
    \begin{tabular}{lccccccc}
    \hline
        Model & B-1 & B-2 & B-3 & B-4 & METEOR & ROUGE-L & CIDEr \\ \hline
        \multicolumn{8}{c}{(a) Scores of image captioning on the MS COCO dataset} \\ 
        Adaptive & $\mathbf{0.743}$ & $\mathbf{0.572}$ & 0.424 & 0.313 & 0.265 & 0.546 & 1.088 \\ 
        Vanilla-ED & 0.729 & 0.556 & 0.409 & 0.299 & 0.249 & 0.531 & 0.952 \\ 
        Attr-Only & 0.671 & 0.495 & 0.366 & 0.276 & 0.255 & 0.535 & 1.059 \\
        \aicab & 0.730 & 0.554 & $\mathbf{0.424}$ & $\mathbf{0.339}$ & $\mathbf{0.279}$ & $\mathbf{0.550}$ & $\mathbf{1.105}$ \\ 
        \multicolumn{8}{c}{(b) Scores of image captioning on the Fashion Bestseller dataset} \\ 
        Adaptive & $0.365$ & $0.306$ & $0.275$ & $0.255$ & $0.185$ & $0.334$ & $2.094$ \\
        Vanilla-ED & $0.345$ & $0.284$ & $0.251$ & $0.231$ & $0.173$ & $0.316$ & $1.870$ \\
        Attr-Only & $0.350$ & $0.290$ & $0.258$ & $0.240$ & $0.179$ & $0.321$ & $1.988$ \\
        \aicab & $\mathbf{0.385}$ & $\mathbf{0.316}$ & $\mathbf{0.289}$ & $\mathbf{0.280}$ & $\mathbf{0.190}$ & $\mathbf{0.349}$ & $\mathbf{2.189}$ \\
        \multicolumn{8}{c}{(c) Scores of image captioning on the Fashion 9 vendors dataset} \\ 
        Adaptive & $0.276$ & $0.218$ & $0.193$ & $0.178$ & $0.121$ & $0.256$ & $1.202$ \\
        Vanilla-ED & $0.264$ & $0.216$ & $0.179$ & $0.165$ & $0.115$ & $0.238$ & $1.149$ \\
        Attr-Only & $0.268$ & $0.210$ & $0.184$ & $0.169$ & $0.117$ & $0.242$ & $1.155$ \\
        \aicab & $\mathbf{0.290}$ & $\mathbf{0.231}$ & $\mathbf{0.202}$ & $\mathbf{0.191}$ & $\mathbf{0.125}$ & $\mathbf{0.268}$ & $\mathbf{1.297}$ \\
    \hline
    \end{tabular}
\end{table*}

We evaluate image captioning on the MS COCO dataset, the Fashion Bestseller dataset, and the Fashion 9 vendors dataset. Table~\ref{tab:image_captioning_all_three} reports the evaluation results for these three datasets, where B-$n$ is BLEU score that uses up to $n$-grams. In each column, higher is better.

We observe that our \aicab achieves the best performance compared to all three ablated versions. The ablation study reveals the complementarity of all constituents of \aicab.
In terms of CIDEr score, the Vanilla Encoder-Decoder network underperforms \aicab by 0.143, 0.319, and 0.148; the Adaptive attention network by 0.017, 0.095, and 0.095; the Attributes-combined model by 0.107, 0.201, and 0.142.
Note that adaptive attention architecture improves the performance better than the attribute information. These results indicate that two components indeed complement each other, and their co-existence crucially benefits the caption generation.

The three experimental conditions establish a comprehensive spectrum. The general image dataset, MS COCO, 
is the most complex and contains multi objects in each image, for which the CIDEr score is the lowest across the datasets. We only compare the CIDEr score because it is the only metric that keeps a stable scale when the number of captions varies. The fashion dataset contains one single object per image. The Fashion Bestseller dataset is simpler than the Fashion 9 vendors dataset. Although the effectiveness of our network is still obvious, the performance gap widens as the task gets more complicated.

Although the attributes-combined model obtains a similar CIDEr score on the COCO dataset compared with the adaptive attention model, it observably underperforms on other scores. CIDEr focuses more on semantical correctness, while others reflect on grammaticality correctness~\cite{Hossain2019}. These results indicate that attribute information provides significant semantic information. However, to demonstrate these attributes in the generated captions, the model achieves this at the expense of grammatical correctness as a result on MS COCO shows ``attr-only'' gains a similar CIDEr score as ``Adaptive'' but its BLEU scores are significantly poorer. Interestingly, it does not happen on the Fashion dataset. We argue that this is because of the small number of captions. The sentence pattern is easier to recognize on the Fashion dataset. However, this effect is not shown in our \aicab network. It reveals the attention architecture, especially the sentinel gate, corrects the bias brought by the attributes. The two components indeed complement each other.

On the fashion dataset, we observe that our model achieves better performance on the Fashion Bestseller dataset than the Fashion 9 vendors dataset, with an improvement of 0.892 (CIDEr score). This observation is the opposite of the regular pattern in which the increased data size improves the ML model's performance. The reason is the distinct styles and forms of captions from different vendors. The huge gaps in captions from one vendor to another are caused by the sub-standard labeling of the Fashion 9 vendors dataset.

\subsection{Attention Distribution Analysis}\label{attention_distribution_analysis}

\begin{figure*}
  \centering
    \includegraphics[width=\textwidth]{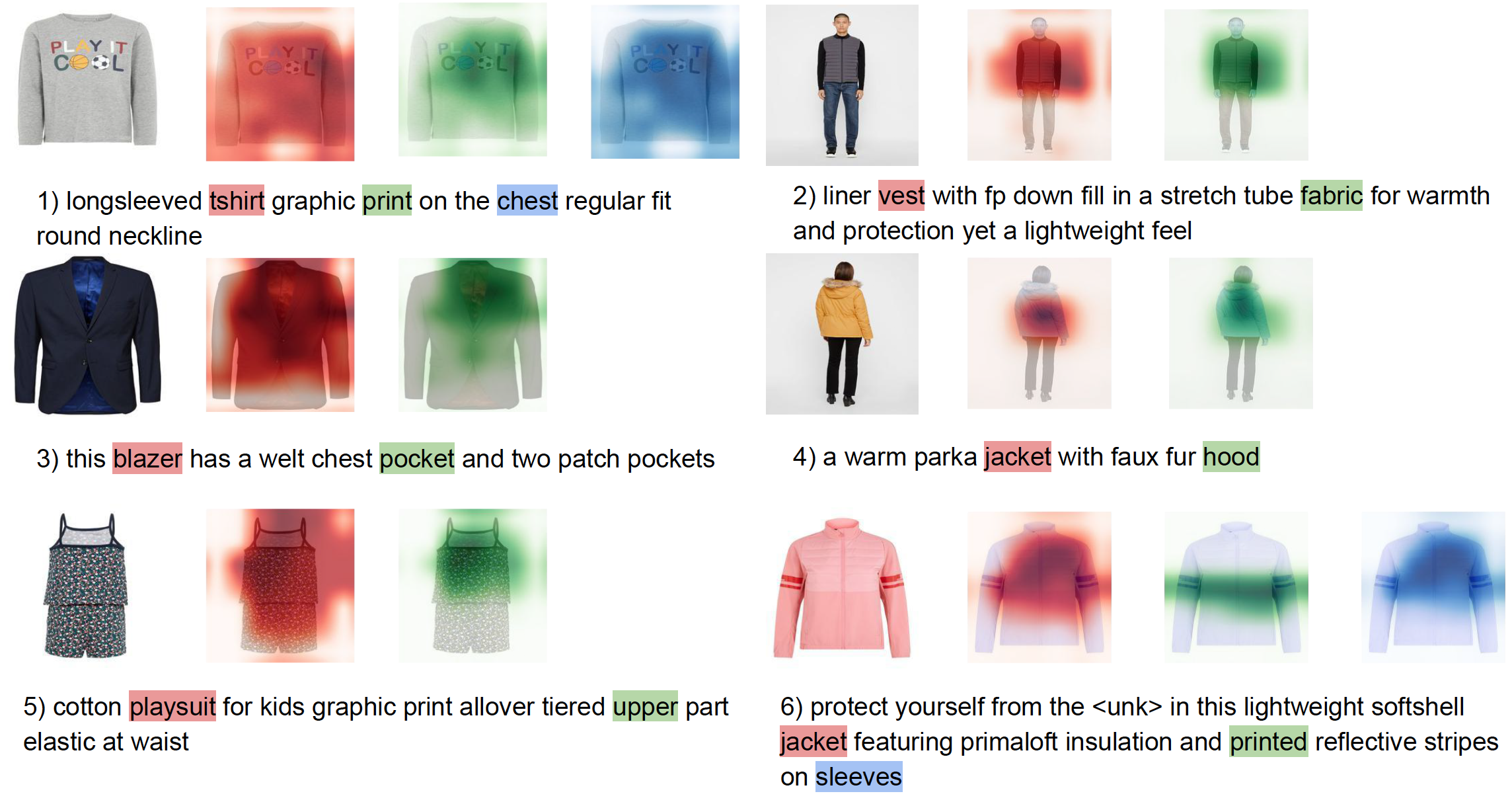}
  \caption{\label{fig:attention_distribution}Visualization of generated captions and image attention maps on the Fashion Bestseller dataset. The text is the captions generated by \aicab. Different colors denote a correspondence between masked regions and underlined words. The first 5 cases are success cases; the last case is a failure.}
\end{figure*}

To better understand our model, we also visualize the image attention distributions $\alpha$ for the generated caption. Using bilinear interpolation and pyramid expansion, we sample the attention map to the image size ($224\times224$). Fig.~\ref{fig:attention_distribution} shows the generated captions and the image attention distribution for specific words in the caption. The first five cases are success cases, and the last case shows a failure example. We see that our model learns to 
pay attention to the specific region when generating different words in the caption, which corresponds strongly with human intuition. Note that on the failure case, although our model fails to focus on the region of the sleeves when generating ``sleeves'', it still successfully recognizes the position of the printed stripe. 

\begin{figure*}
\begin{subfigure}[b]{0.6\textwidth}
\centering
\includegraphics[scale=0.25]{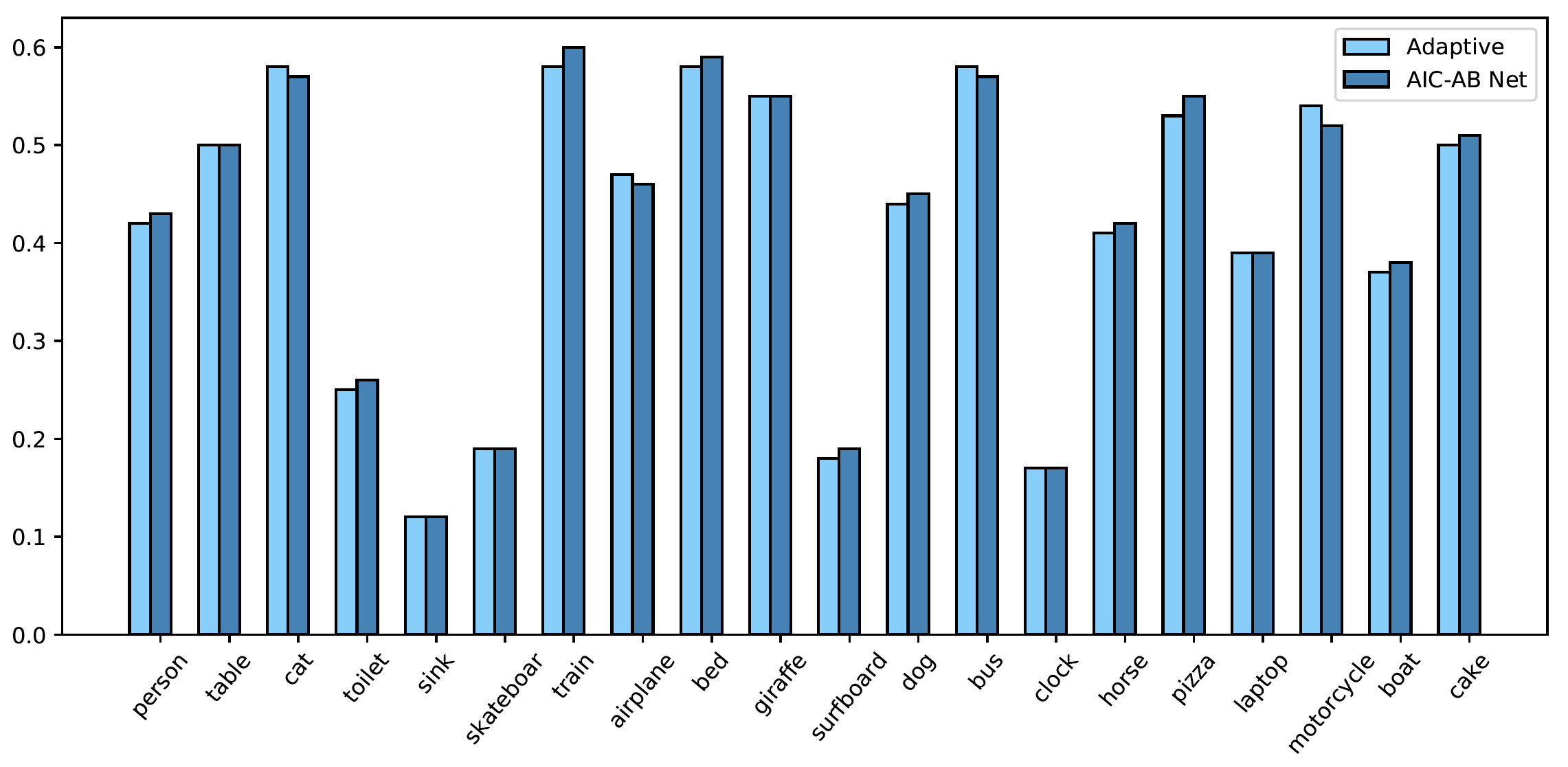}
\caption{For 20 most frequent COCO object categories.}
\label{fig:localization}
\end{subfigure}
\hfill
\begin{subfigure}[b]{0.4\textwidth}
\centering
\includegraphics[scale=0.20]{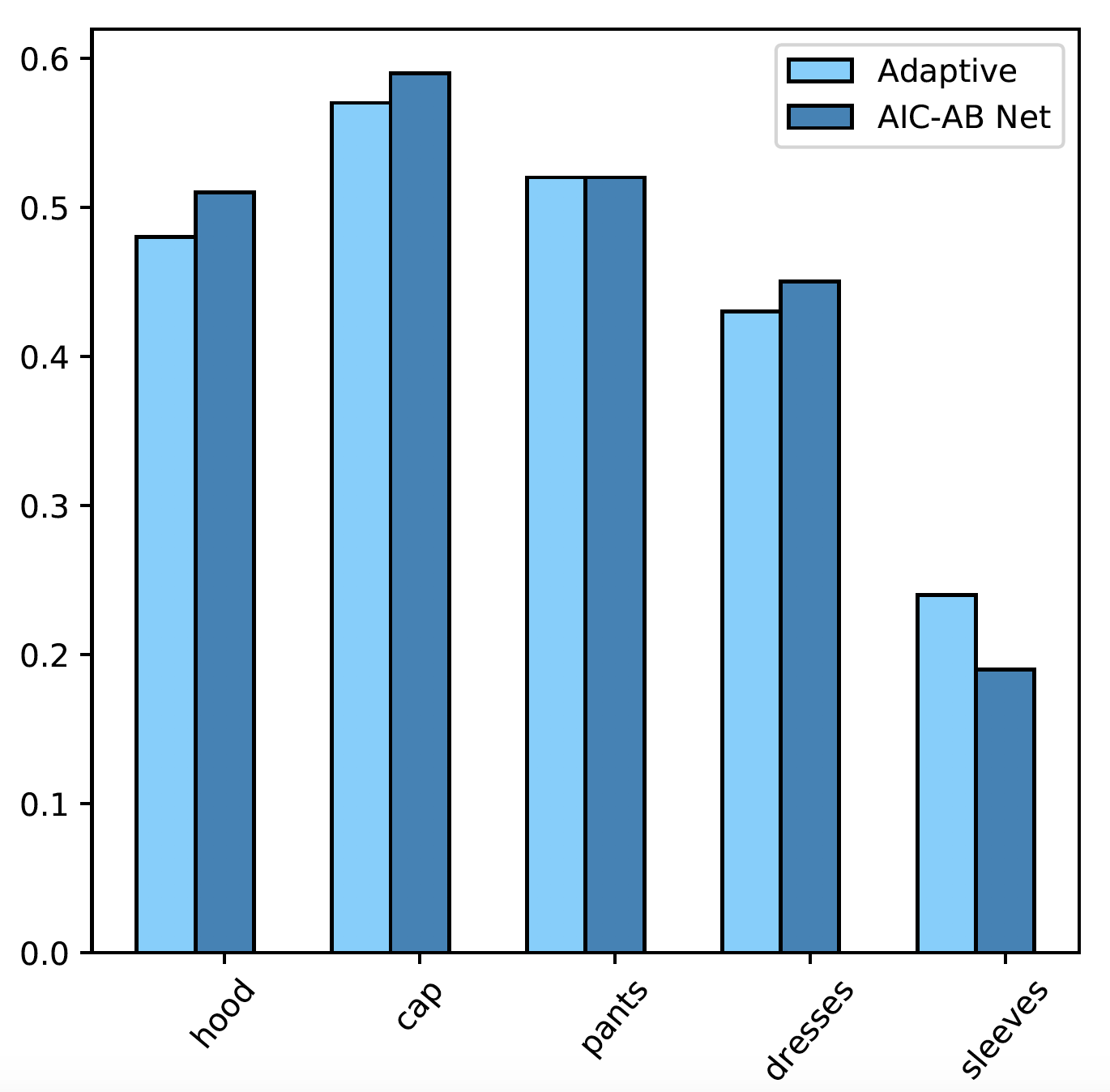}
\caption{For three typical words.}
\label{fig:fashion_localization}
\end{subfigure}
\caption{Localization accuracy over generated captions. Adaptive is the baseline model (an ablated version of \aicab); \aicab is our model.}
\end{figure*}

Since the COCO dataset provides the ground truth of objects’ bounding box, it can be used to evaluate the performance of attention map generation. The spatial intersection over union (sIOU) score is used to measure localization accuracy. Given the word $w_t$ and its corresponding attention map ${\alpha}_t$, we first segment the regions of the image with its attention value more extensive than a pre-class threshold $th$ (after the map is standardly normalized to scale [0,1), where we set as $0.6$. Then we take the bounding box covering the largest connected component in this segmentation map as the predicted attention region. We report the sIOU between the predicted bounding box and the ground truth for the top 20 most frequent COCO object categories, as Fig.~\ref{fig:localization} shows. The average localization accuracy for the ``Adaptive'' is 0.415, and 0.419 for our \aicab. This implies that the attribute information benefits the attention map generation as a combined model. We also observe that our \aicab and its attention-only version have a similar trend. They both perform well on informative visual objects and large objects such as ``cat'', ``train'', ``bed'', and ``bus'', while they have poor performance on small objects such as ``sink'' and ``clock''. We argue that it is because our attention map is extracted from $7 \times 7$ spatial map, which loses plenty of resolution and detail. This defect is remarkably exposed when detecting small objects. This reason can explain the wrong attention map on the Fashion dataset as well, where the majority of words are descriptions of details and refer to small regions on the image.

Since the bounding box ground truth is missing in the Fashion dataset, we apply statistic analysis for 5 typical words as quantitative analysis, \textit{hood}, \textit{cap}, \textit{pants}, \textit{dresses}, and \textit{sleeves}. In common cases, \textit{hood} and \textit{cap} only show on the upper part of an image, \textit{pants} and \textit{dresses} only on the lower part, and \textit{sleeves} only on the left and right sides. We assume these regions are their ground truth respectively and apply the same approach as explained above to measure the localization accuracy. Fig.~\ref{fig:fashion_localization} reports the result of the Fashion dataset. We observe that \aicab outperforms on the first 4 words than the word \textit{sleeves} and shows a similar trend with the adaptive attention model.

\section{Conclusion}\label{sec:conclusions}

This work has been motivated by the task of generating captions for single-object fashion images and inspired by the adaptive attention architecture~\cite{Lu2017} and semantic concept~\cite{ting2016}. Toward this end, we present Attribute-Image-Combined Attention-Based Network (\aicab). We have evaluated \aicab on the MS COCO and the Fashion datasets. Our experiments indicate that the ability to locate the relevant region of an image when generating different words and the combination with the attribute information is crucial for accurate caption generation. 

Further research could explore two directions. First, according to the results of transfer learning, we suggest creating an up-to-standard labeling system for the Fashion dataset, which will benefit the consistency of the data and the robustness of the models trained on it. Second, we argue that segmenting the images into more regions will improve the performance since, in some cases, our model cannot pay attention to the accurate region when generating words.

Image captioning is a challenging and promising task for the Internet industry and computer vision. We believe this work represents a significant step in improving image captioning and breeds useful applications in other domains. 
\bibliography{main}
\bibliographystyle{IEEEtran} 
\end{document}